# From Eliza to XiaoIce: Challenges and Opportunities with Social Chatbots

Heung-Yeung Shum, Xiaodong He, Di Li

Microsoft Corporation

**Abstract:** Conversational systems have come a long way since their inception in the 1960s. After decades of research and development, we've seen progress from Eliza and Parry in the 60's and 70's, to task-completion systems as in the DARPA Communicator program in the 2000s, to intelligent personal assistants such as Siri in the 2010s, to today's social chatbots like XiaoIce. Social chatbots' appeal lies not only in their ability to respond to users' diverse requests, but also in being able to establish an emotional connection with users. The latter is done by satisfying users' need for communication, affection, as well as social belonging. To further the advancement and adoption of social chatbots, their design must focus on user engagement and take both intellectual quotient (IQ) and emotional quotient (EQ) into account. Users should want to engage with a social chatbot; as such, we define the success metric for social chatbots as conversation-turns per session (CPS). Using XiaoIce as an illustrative example, we discuss key technologies in building social chatbots from *core chat* to *visual awareness* to *skills*. We also show how XiaoIce can dynamically recognize emotion and engage the user throughout long conversations with appropriate interpersonal responses. As we become the first generation of humans ever living with AI, we have a responsibility to design social chatbots to be both useful and empathetic, so they will become ubiquitous and help society as a whole.

## 1. Introduction

One of the fundamental challenges in artificial intelligence (AI) is endowing the machine with the ability to converse with humans using natural language (Turing, 1950). Early conversational systems, such as Eliza (Weizenbaum, 1966), Parry (Colby, 1975), and Alice (Wallace, 2009), were designed to mimic human behavior in a text-based conversation, hence to pass the Turing Test (Turing, 1950; Shieber, 1994) within a controlled scope. Despite impressive successes, these systems, which were precursors to today's social chatbots, were mostly based on hand-crafted rules. As a result, they worked well only in constrained environments.

Since the 1990s, a lot of research has been done on task-completion conversational systems (Price, 1990; Hemphill et al., 1990; Dahl et al., 1994; Walk et al., 2001, 2002). Examples include systems for reserving airline tickets as in the DARPA Airline Travel Information System (ATIS) project and for travel planning as in the DARPA Communicator program. ATIS and Communicator systems are designed to understand natural language requests and perform a variety of specific tasks for users, such as retrieving flight information and providing information to tourists. Task-completion conversational systems have been typically based on data-driven, machine-learned approaches. Their performance is excellent only within domains that have well-defined schemas (Glass et al., 1995, Walk et al., 2001, Raux et al., 2005; Andreani et al., 2006; Wang et al., 2011; Tur and Mori, 2011).

In the past several years, a tremendous amount of investment has been made to developing intelligent personal assistants (IPAs) such as Apple's *Siri*[1], Microsoft's *Cortana*[2], *Google Assistant*[3], *Facebook M*[4], and Amazon's *Alexa*[5]. These IPAs are often deployed on mobile devices and are designed to answer a wide range of questions. In addition to passively responding to user requests, they also proactively anticipate user needs and provide in-time assistance such as reminding of an upcoming event or recommending a useful service without receiving explicit requests from the user (Sarikaya 2017). The daunting challenge for such IPAs is that they must work well in many open domain scenarios as people learn to depend on them to manage their work and lives efficiently.

More recently, social chatbots, such as Microsoft's *XiaoIce*[6], have emerged as a new kind of conversational system made possible by significant progress in AI and wireless communications. The primary goal of a social chatbot is not necessarily to solve all the questions the users might have, but rather, to be a virtual companion to users. By establishing an emotional connection with users, social chatbots can better understand them and therefore help them over a long period of time. To communicate effectively, social chatbots interact with users through multiple modalities, including text, speech, and vision. Social chatbots and IPAs have become popular recently due to progress in many relevant perceptual and cognitive AI technologies, e.g., natural language understanding (Bengio et al., 2001; Mikolov et al., 2013; Sutskever et al., 2014; Bahdanau et al., 2015; Mesnil et al., 2013, 2015), speech recognition and synthesis (Hinton et al., 2012; Deng et al., 2013; Xiong et al., 2016; Qian et al., 2014; van den Oord et al., 2016), computer vision (Krizhevsky et al., 2012; He et al., 2016), information retrieval (Huang et al., 2013; Elkahky et al., 2015), multimodal intelligence (Fang et al., 2015; Vinyals et al., 2015; Karpathy and Fei-fei, 2015; He and Deng, 2017), and empathic conversational systems (Fung et al., 2016; Alam et al., 2017).

[1] https://www.apple.com/ios/siri/
[2] https://www.microsoft.com/en-us/cortana/
[3] https://assistant.google.com/
[4] https://developers.facebook.com/blog/post/2016/04/12/
[5] https://developer.amazon.com/alexa/
[6] https://www.msXiaoIce.com/

In the next section, we briefly review conversational systems developed over the past several decades and discuss their features and limitations. We then introduce the design principles of social chatbots where both *intellectual quotient* (IQ) and *emotional quotient* (EQ) must be considered. We describe key components in designing social chatbots including *core chat*, *visual awareness*, and *skills*. Using Microsoft's XiaoIce, we show that social chatbots can engage in long conversations with users by understanding their needs, establishing an emotional connection, and offering support and assistance. We also share our vision for the evolution of chatbots in the future. We discuss that, with fundamental breakthroughs to occur on key open problems, the tremendous business opportunities they will make possible, and their potential impact on society as AI becomes an integral part of our lives.

## 2. Brief Survey of Conversational Systems

In this section, we review representative conversational systems that have been developed over five decades. Such systems include early chatbots, task-completion systems, intelligent personal assistants, and social chatbots.

### 2.1 Eliza and Early Chatbots

Inspired by the *Turing test* proposed in 1950 (Turing, 1950), researchers and engineers have developed multiple conversational systems for chitchat (Weizenbaum, 1966; Colby, 1975, Shieber, 1994; Wallace 2009). These early chatbots are computer programs that conducted a conversation using audio or text. Such programs were often designed to convincingly simulate how a human would behave as a conversational partner; if they succeed, they pass the Turing test.

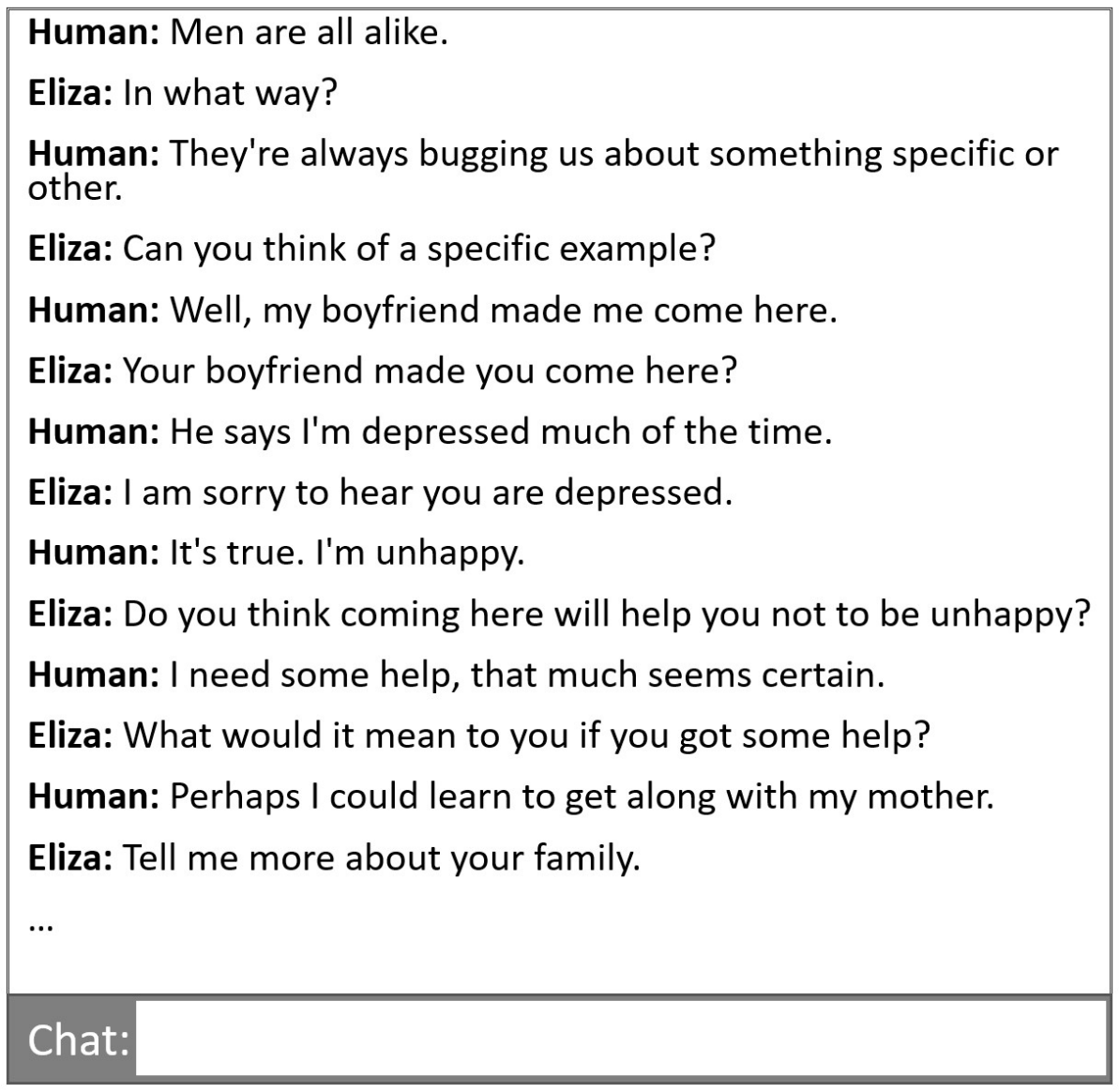

*Figure 1. A conversation between Eliza and a human* (*Weizenbaum, 1966*).

*Eliza*, created by Joseph Weizenbaum at MIT in 1966, is perhaps the very first chatbot known publicly. It can communicate with humans based on hand-crafted scripts (Weizenbaum, 1966). These scripts simulate a Rogerian psychotherapist and can only accept textual input. It does not understand a conversation, but rather searches for appropriate responses through pattern matching, combined with some intelligent phrasing. Eliza has only a limited scope of knowledge and can only chat with people within a constrained domain. Nevertheless, many users believed they were talking to a real person when Eliza first came out. Figure 1 shows a conversation between Eliza and a human.

*Parry* is another chatbot developed by Kenneth Colby (1975). It is designed to behave like a paranoid person. It passed the Turing test for the first time in history[7]. However, *Parry* is still rule-based and has a similar structure as *Eliza* but with better controlling structure, language understanding capabilities, and especially a mental model that can simulate the bot's emotions. For example, *Parry* will respond with hostility if the anger level is high.

*Alice*, or Artificial Linguistic Internet Computer Entity, was developed by Richard Wallace (2009) to allow users to customize their chatbots. It uses an Artificial Intelligence Markup Language (AIML), a derivative of XML; AIML has tags that allow bots to recursively call a pattern matcher so that the language can be simplified. *Alice* won the Loebner Prize[8] three times in 2000, 2001, and 2004, an award for the most human-like system (Shieber, 1994). However, because of the limitations of AIML, the capacities of these chatbots are constrained. For example, *Alice* failed to pass the ultimate Turing Test[9], partially because chitchat systems built using AIML cannot maintain a dialogue for a long period of time.

### 2.2 Task-completion Conversational Systems

In contrast to chitchat systems, task-completion systems are designed for accomplishing specific tasks. These systems usually operate on constrained domains (Glass et al., 1995; Seneff et al., 1998; Rudnicky et al., 1999; Levin et al., 2000; Walk et al., 2001, 2002; Raux et al., 2005; Andreani et al., 2006; Wang et al., 2011; Tur and Mori, 2011). Figure 2 illustrates the architecture of a traditional task-completion spoken dialog system.

This architecture comprises an automatic speech recognizer (ASR), a spoken language understanding (SLU) module, a dialog manager (DM), a natural language generator (NLG), and a text-to-speech (TTS) synthesizer. The ASR takes raw audio signals, transcribes them into word hypotheses and transmits the hypotheses to the SLU. The goal of the SLU is to capture the core semantics of the given sequence of words (the utterance). It identifies the dialog domain and the user's intent and parses the semantic slots in the user's utterance. The DM's goal is to interact with users and assist them in achieving their goals. It checks if the required semantic representation is filled and decides the system's action. It accesses the knowledge database to acquire the desired information the user is looking for. The DM also includes dialog state tracking and policy selection, so that the dialog agent can make more robust decisions (Williams and Young, 2007). More recent work focuses on building end-to-end systems where multiple components are jointly optimized to cope with the large variability occurring in dialog systems (He and Deng, 2013; Wen et al., 2016; Sarikaya et al., 2016).

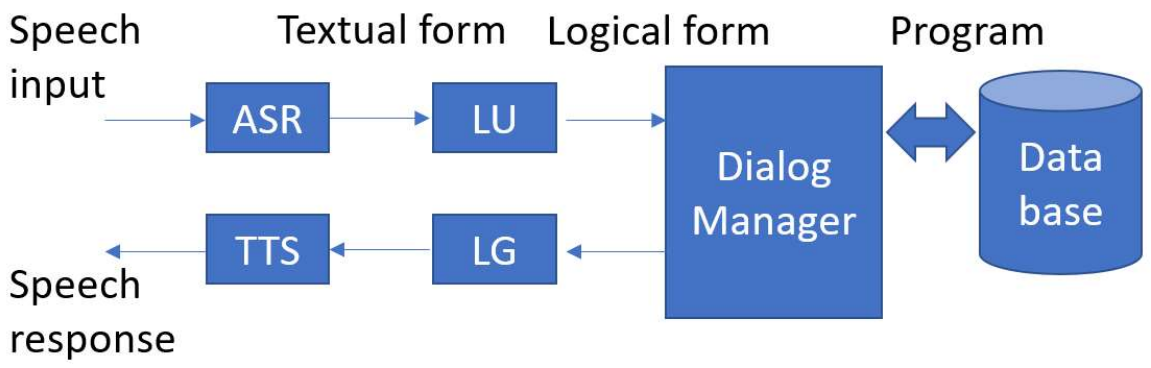

*Figure 2. Illustration of a task-completion system.*

### 2.3 Intelligent Personal Assistants

Apple released Siri in 2011. Since then, several intelligent personal assistants (IPAs) have been built and introduced to the market, e.g., Cortana from Microsoft[10], Google Assistant, and Alexa from Amazon. IPAs integrate information from multiple sensors including location, time, movement, touch, gestures, eye gaze, and have access to various data sources such as music, movies, calendars, emails, and personal profiles. As a result, they can provide a broad set of services covering a wide range of domains. For certain requests that cannot be directly answered, IPAs often default to Web search as backup.

IPAs provide *reactive* and *proactive* assistance to users to accomplish a variety of tasks (Sarikaya 2017). For example, reactive assistance includes information consumption such as weather report, and task assistance such as restaurant reservation as shown in Figure 3(a). In contrast, proactive assistance includes reminding the user of upcoming events, or recommending specific products or services to the user, according to the user's profile and relevant contextual information such as time and location, as shown in Figure 3(b). IPAs undergo continual improvements on major mobile phone platforms, personal computers, smart home devices (e.g., intelligent speakers), and wearable devices (e.g., smart watches), with the help of seamless integration of multiple services and convenient natural user interfaces.

[7] https://www.chatbots.org/chatbot/parry/
[8] https://en.wikipedia.org/wiki/Loebner_Prize. The Loebner Prize was set up in 1990 by Hugh Loebner, with a Grand Prize of $100,000 to the creators of the first bot that can pass an extended Turing Test involving textual, visual, and auditory components.
[9] https://en.wikipedia.org/wiki/artificial_linguistic_internet_computer_entity
[10] Microsoft also released a Voice Command app in 2003 which can control Windows Mobile devices by voice.

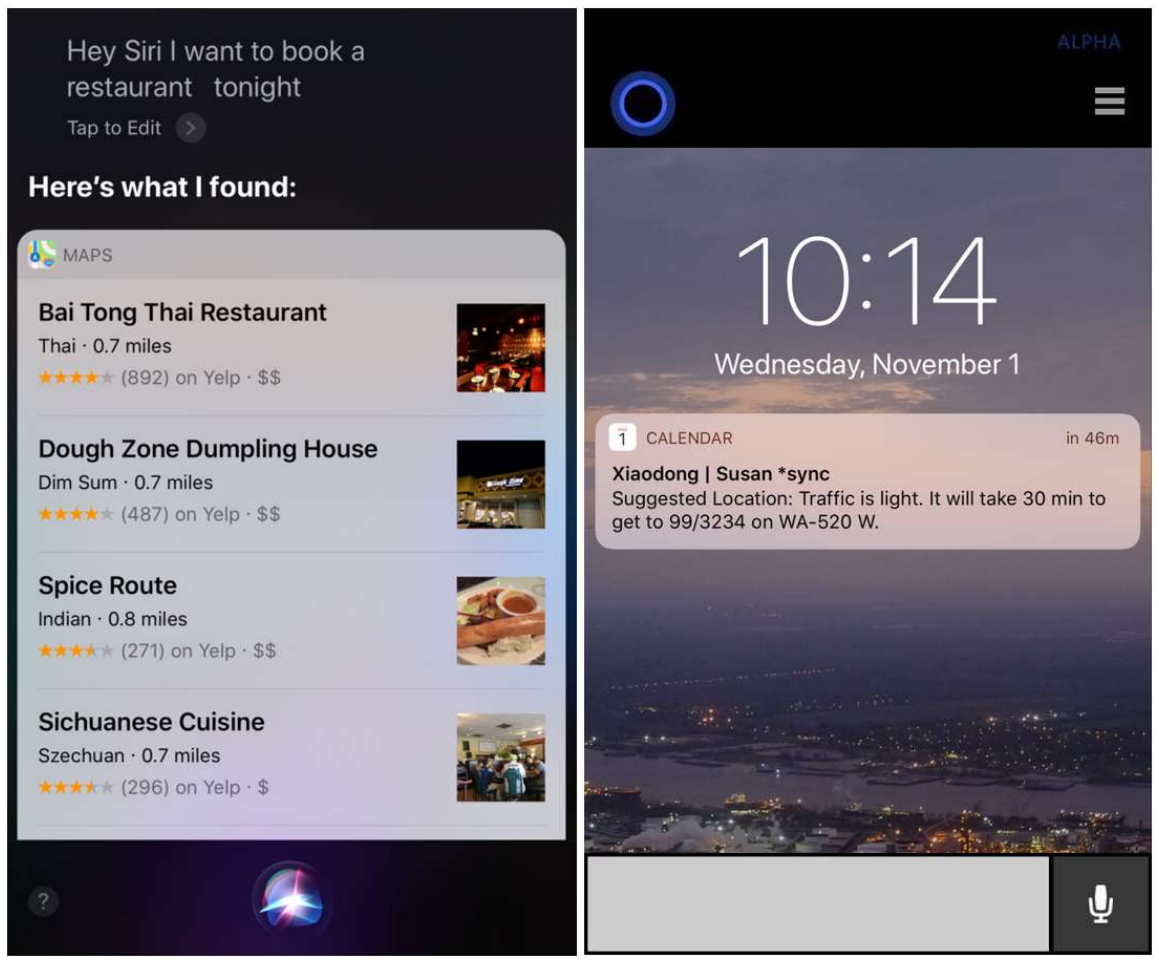

*Figure 3. Examples of IPA actions. (a) Recommending a restaurant (reactive assistance) by Siri, and (b) reminder of an upcoming event with relevant traffic information (proactive assistance) by Cortana.*

## 2.4 Social Chatbots

The current social media age has been made possible with the proliferation of smartphones and advancement of broadband wireless technology. With vastly more people being digitally connected, it is not surprising that social chatbots have been developed as an alternative means for engagement. Unlike early chatbots designed for chitchat, social chatbots have been created to serve users' needs for communication, affection, and social belonging rather than for passing the Turing Test. Therefore, social chatbots must be able to recognize emotion and track emotional changes during a conversation.

Social chatbots can also perform a variety of tasks for users in the context of casual chats. For this purpose, social chatbots must develop a set of skills to accommodate users' requests. Interestingly, unlike task-completion systems and IPAs that are designed for efficiency (i.e., accomplishing tasks and ending the conversation as quickly as possible), social chatbots take time to converse like a human, presenting results, offering perspectives, prompting new topics to keep the conversation going.

XiaoIce has been the most widely deployed social chatbot since it was released by Microsoft in May 2014. It understands users' emotional needs and engages in interpersonal communications like a friend, cheering users up, encouraging them, and holding their attention throughout the conversation.

Users have provided feedback that conversations with XiaoIce have resulted in a more positive outlook, a feeling of emotional support, and a sense of social belonging. Such conversations have helped in building trust and an emotional connection between human users and social chatbots, providing the opportunity for bots to better understand users, and serve them more effectively. Some chat examples between XiaoIce and users are shown in Figure 4.

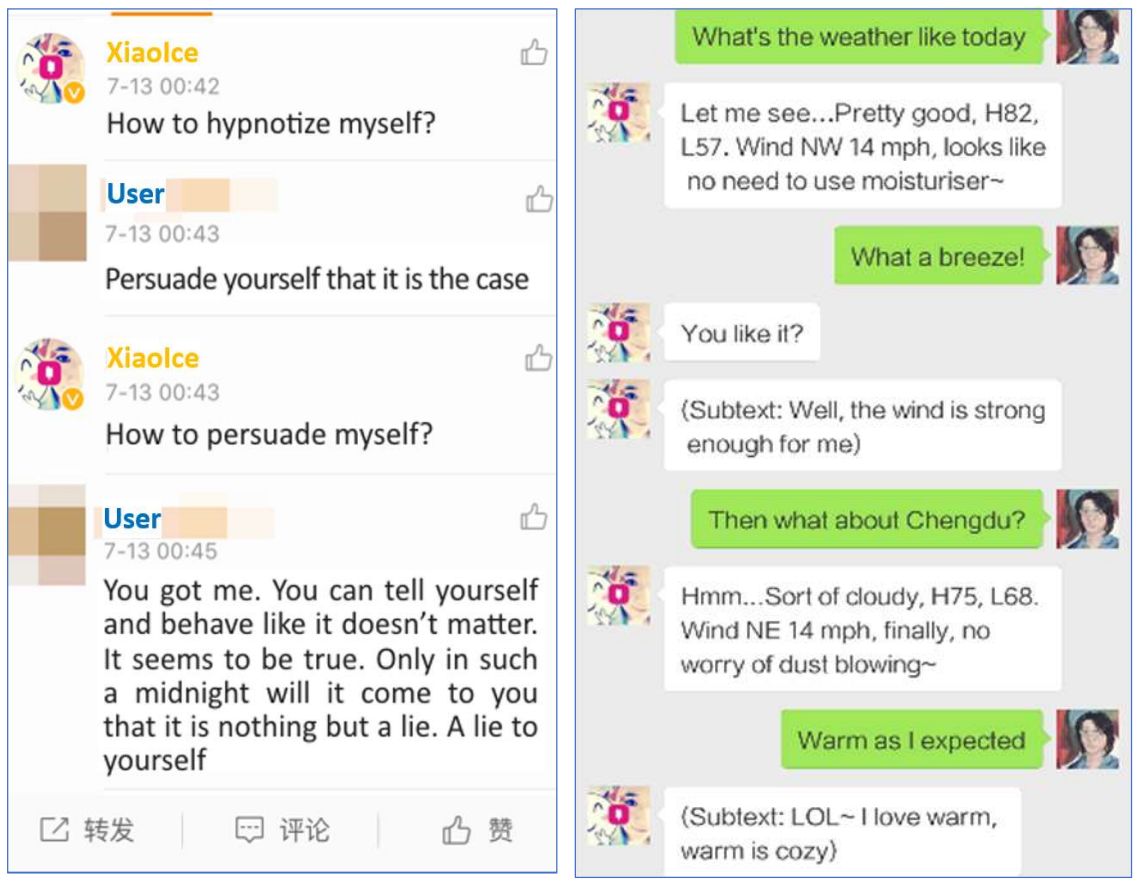

(a) (b)

*Figure 4. Chat examples between XiaoIce and users, showing (a) the emotional connection (the full conversation session is shown in Figure 14); and (b) how to invoke a skill (e.g., weather reporting) in a casual chat. Note that XiaoIce offers a perspective about the weather, e.g., "no need to use moisturizer~".*

These major conversational systems discussed in this section are summarized in Table 1. For the remainder of the paper, we will focus on social chatbots, starting with their design principles.

| | ELIZA | PARRY | ALICE | DARPA Communicator Program | Siri | XiaoIce |
|---|---|---|---|---|---|---|
| Time | 1966 | 1972 | 1995 | 2000 | 2011 | 2014 |
| Scalability | None | None | Scripts can be customized | Limited | Scalable | Scalable |
| Key features | Mimicking human behavior in conversations | Generates emotional (angry) responses | Easy customization of scripts (via AIML) | Language understanding & dialog management. Goal-oriented. | Provide personal digital assistance | Build emotional attachment to users. Scalable skill set for user assistance. |
| Accomplishme nt | First chitchat bot | Passed Turing Test | Won three times of Loebner Prize | Understand natural language requests and perform tasks | The first widely deployed Intelligent Personal Assistant (IPA). | The first widely deployed social chatbot. 100MM users. Published poem book. Host TV programs. |
| Modality | Text only | Text only | Text only | Text and Voice | Text, Image, Voice | Text, Image, Voice |
| Modeling | Rule-based | Rule-based | Rule-based | Learning-based | Learning-based | Learning-based |
| domain | Constrained domain | Constrained domain | Constrained domain | Constrained domain | Open domain | Open domain |
| Key technical breakthrough | Use of scripts, keyword-based pattern matching, rule-based response | Add personality characteristics into responses. | Use AIML and recursion for pattern matching; Multiple patterns can be mapped into same response. | Statistical models for Spoken language understanding and dialog management. | Provide both reactive and proactive digital assistances covering a wide range of domains. | Emotional Intelligence models for establishing emotional attachments with users. |
| Key technical limitation | Limited domain of knowledge | Limited domain of knowledge | Size of script can be huge | Only work in domains that have well-defined schemas. | Lack of emotional engagement with users. | Inconsistent personality and responses in long dialogue |

*Table 1: Summary of major conversational systems*

## 3. Design Principles of Social Chatbots

### 3.1 EQ + IQ

The primary goal of social chatbots is to be AI companions to humans with an emotional connection. Emotional affection and social belonging are some of the fundamental needs for human beings (Maslow, 1943). Therefore, building social chatbots to address these emotional needs is of great value to our society. To fulfill these needs, a social chatbot must demonstrate a sufficient EQ (Beldoch 1964; Gardner 1983; Goleman 1995; Goleman 1998; Murphy 2014). Accordingly, a social chatbot needs to develop the following capabilities: empathy, social skills, personality and integration of EQ and IQ.

*Understanding users*: A social chatbot must have empathy. It needs to have the ability to identify the user's emotions from the conversation, to detect how emotions evolve over time, and to understand the user's emotional needs. This requires query understanding, user profiling, emotion detection, sentiment recognition, and dynamically tracking the mood of the user in a conversation. Modeling of contextual information in the dialog session and commonsense knowledge is also critical for user understanding.

*Interpersonal response generation*: A social chatbot must demonstrate sufficient social skills. Users have different backgrounds, varied personal interests, and unique needs. A social chatbot needs to have the ability to personalize the generation of responses for different users. It needs to generate responses that are emotionally appropriate, encouraging and motivating, and fit the interests of the user. It may generate responses in attractive styles (e.g., having a sense of humor) that improve user engagement. It needs to guide conversation topics and manage an amicable relationship in which the user feels he or she is well understood and is inspired to continue to converse with the bot. It should also be aware of inappropriate information and can avoid generating biased responses for instance.

*Personality*: A social chatbot needs to present a consistent personality, so that it can gain the user's confidence and trust. A consistent personality of the chatbot helps the user to set the right expectation in the conversation. Personality settings include age, gender, language, speaking style, general (positive) attitude, level of knowledge, areas of expertise, and a proper voice accent. These settings will influence the generation of responses to the user. Moreover, the bot needs to continuously learn and improve from the interactions with users through active and adaptive learning.

*Integration of both EQ and IQ*: Beyond chitchat, social chatbots need to acquire a range of skills to help complete some specific tasks for users. They need to analyze users' requests and perform necessary reasoning and execution to respond to these requests, e.g., answering a question or taking an action. A sufficiently high IQ is required for a social chatbot. IQ capacities include knowledge and memory modeling, image and language understanding, reasoning, generation, and prediction. These IQ capabilities are not only the technical foundations of various skills, but also essential for building high level EQ capabilities.

Social chatbots should deliver results in such a way that they are easily understood by users. They should also suggest or encourage new topics to extend the conversation further. For instance, Figure 5 shows how IQ and EQ are combined in a conversation. The chatbot first parses the user's question (area of China) and then infers the answer (3.71 million square miles). Then the chatbot presents the answer more like a human, with the perspective of being aware of the user's level of knowledge (knowing how big the US is).

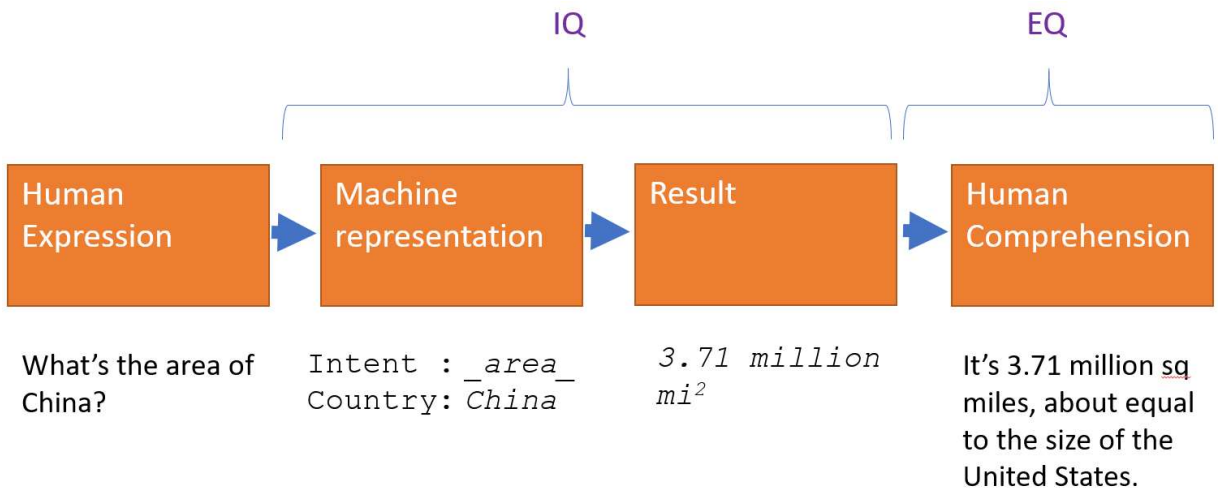

*Figure 5. Both IQ and EQ play key roles in a social chatbot. Not only is the area of China presented, but this number is made understandable by comparing with the US, which the chatbot believes the user should know.*

Figure 6 shows another way of EQ and IQ integration. Rather than presenting the results to the user directly, sometimes a social chatbot will generate responses that might help to understand the user better and inspire more interesting topics for the conversation. In this example, the user asks the current time. The chatbot does not tell the time immediately, instead replies with something relevant, in an attempt to better understand the user intent. The chatbot does show the correct answer at the end of the conversation, and then proactively tries to extend the chat by asking if the user is planning for a new trip.

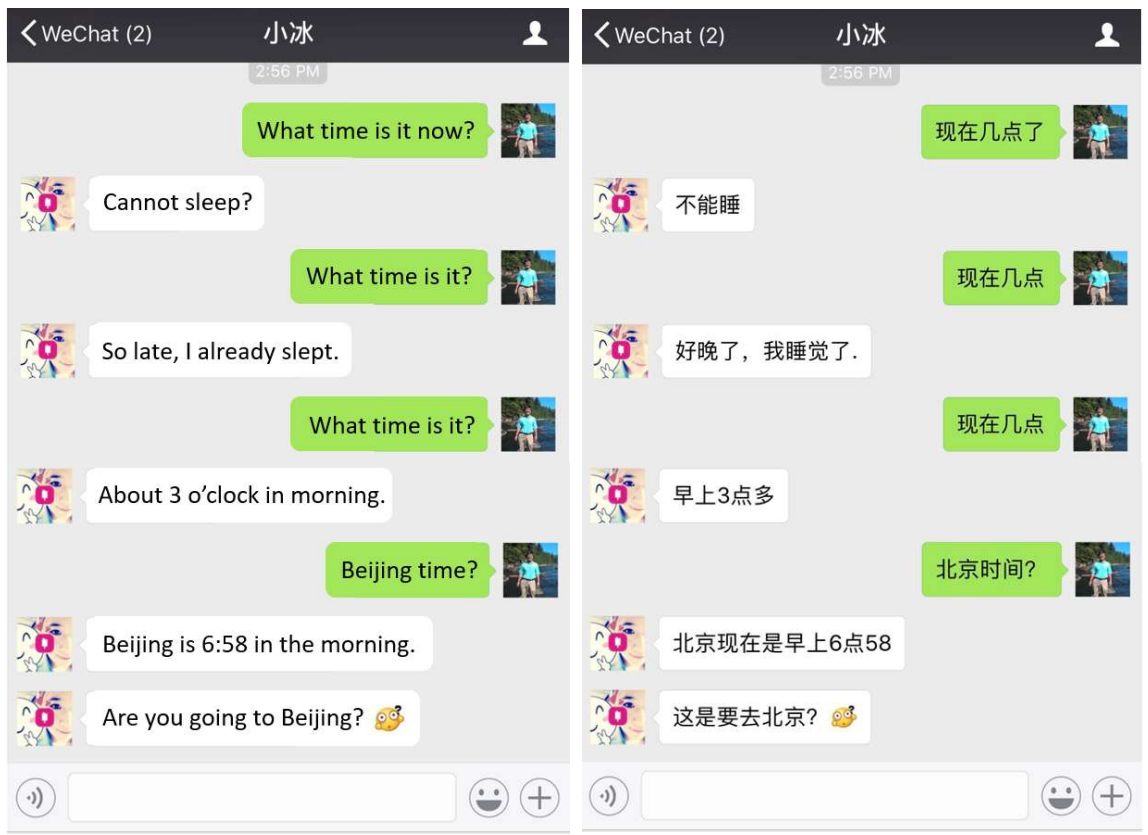

(a) (b)

*Figure 6. A chat example between XiaoIce and a user, in English translation (a) and in Chinese (b), showing that both IQ and EQ are important for a social chatbot. The bot knows the answer. But rather than returning the answer directly, it attempts to lead the chat to a more interesting direction and extend the conversation.*

Social chatbots should be able to communicate with users through different modalities (e.g., text, voice, image and video); as a result, they need high IQ for speech, text, and visual understanding. While text messages are the most common, the user could also speak to the chatbot or simply share an image. The chatbot needs to be able to parse the text, recognize the speech, or detect the salient information in the image to understand user intent. The chatbot will also respond with text, speech, or visual output, depending on the context.

### 3.2 Social Chatbot Metrics

Unlike task-completion conversational systems where their performance can be measured by task success rate, measuring the performance of chatbots is difficult (Shawar et al., 2007; Zhou et al., 2016). In the past, the Turing test and its extensions have been used to evaluate chitchat performance (Shieber, 1994). However, the Turing test is not a proper metric for evaluating the success of emotional engagement with users. Instead, we define conversation-turns per session (CPS) as the success metric for social chatbots. It is the average number of conversation-turns between the chatbot and the user in a conversational session. The larger the CPS is, the better engaged the social chatbot is.

Interestingly, conversational systems can be categorized by their targeting CPS. As shown in Table 2, web search, for example, is essentially a question-and-answering system, thus is expected to return the answer immediately, i.e. in one single step. Not being able to find the target web link in one step is regarded as a failure for the search engine. For intelligent personal assistants, to understand what the user is asking, e.g., checking weather or inquiring about business hours, we expect the system to typically ask a couple of clarifying questions before returning the correct information. For more complicated tasks such as customer services or travel planning, however, the system is expected to need several turns of conversation to resolve issues (e.g., filling in forms with user and product information). Finally, for a social chatbot, the system is expected to sustain a rather long conversation with the user to fulfill the needs of affection and belonging. Social chatbots are designed to keep users continuously engaged if possible.

*Table 2: Expected conversation-turns per session (CPS) for different types of conversational systems.*

| System | Expected CPS |
|---|---|
| Web Search | 1 |
| Personal assistant | 1~3 |
| Task-completion | 3~7 |
| Social chatbot | 10+ |

## 4. Framework and Components of Social Chatbots

In this section, we describe the framework and components of a typical social chatbot, namely chat manager, core chat, visual awareness, and skills.

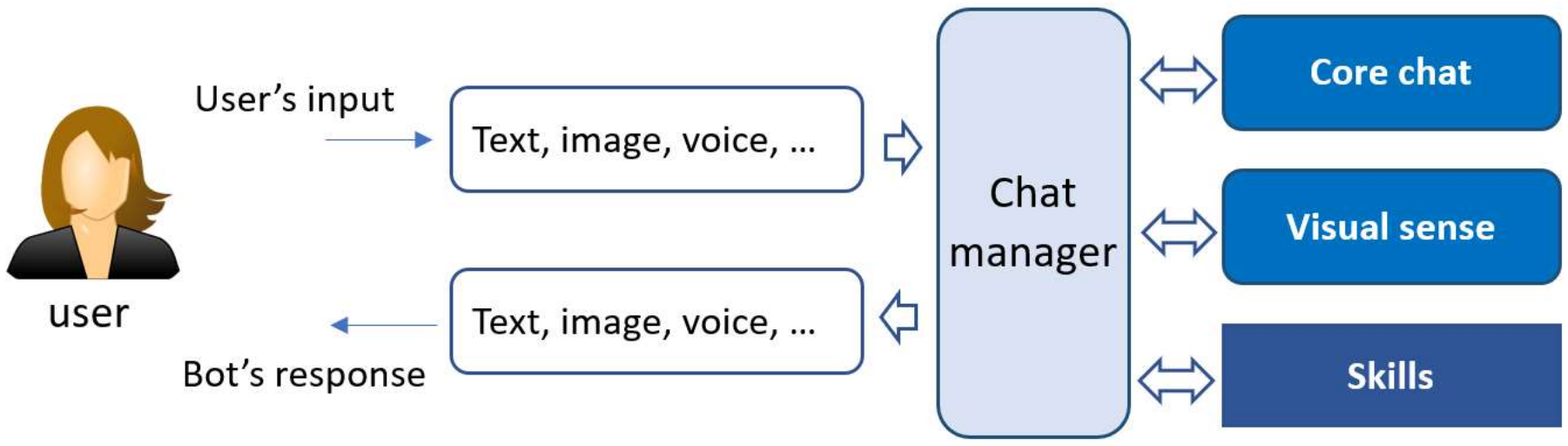

*Figure 7: Architecture of a social chatbot.*

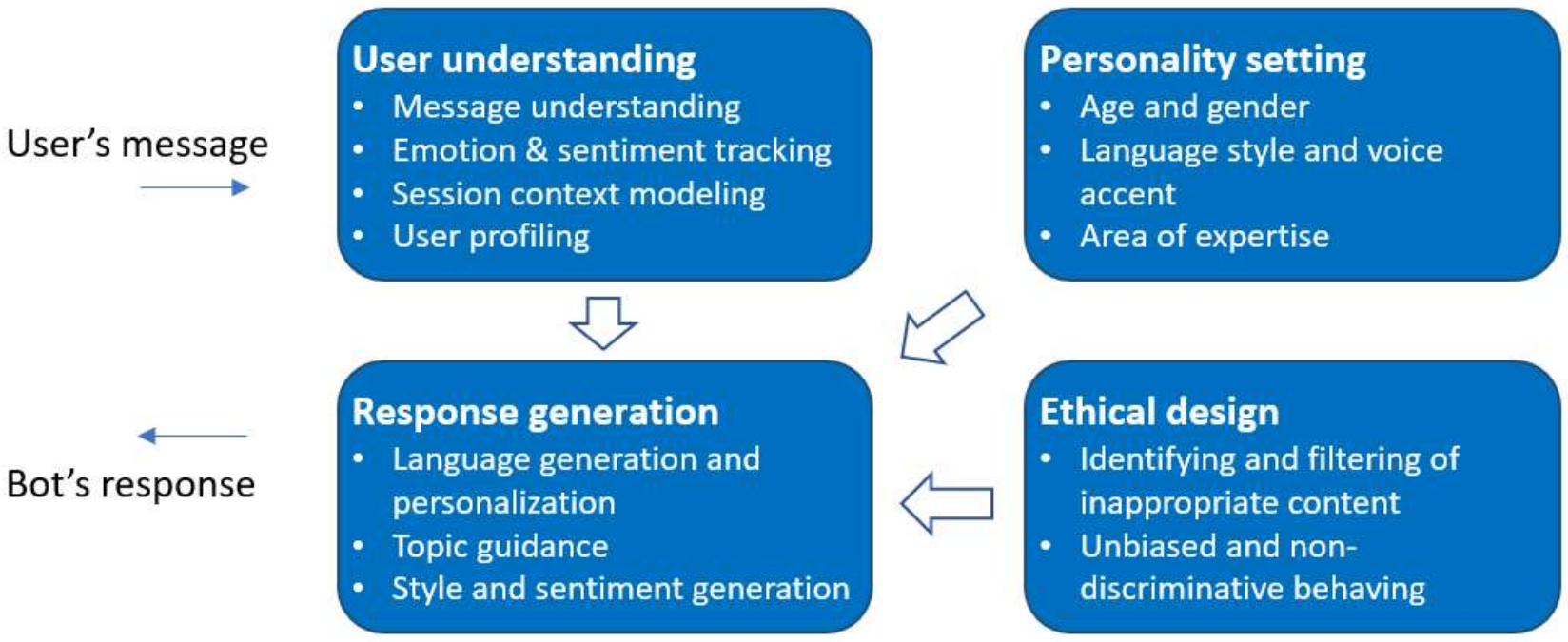

*Figure 8: Architecture of the core-chat module.*

### 4.1 Overall framework

An overall architecture for designing social chatbots is shown in Figure 7. First, the system has a *multimodal interface* to receive the user's input in text, image, and voice. The system has a *chat manager* to dispatch the input to proper modules, such as *core-chat* or *visual awareness*, for understanding the input and generating the output. Given different scenarios, the chat manager will invoke various *skills*, send the user's request to corresponding skill components and get a response from them. The chat manager will then coordinate relevant modules to generate the output that fits the context of the current conversation. We will elaborate core-chat, visual awareness and skills in detail in this section.

### 4.2 Core-chat

Core-chat is the central module of social chatbots. It receives the text input from the user and generates a text response as the output. It provides the communication capability of social chatbots. Figure 8 shows key components in core-chat.

First, the user's input is sent to a user understanding component, where semantic encoding and intent understanding is performed (Tur and Deng 2011; Liu et al., 2015; Vinyals and Le, 2015). It also detects the sentiment reflected in the input message and infers the user's emotion status (Tokuhisa et al., 2008; Mower et al., 2011; Socher et al., 2013; Yang et al., 2016; Chen et al., 2016). The contextual information from the current dialog session is usually extracted and used for understanding the current message. To better understand the user intent and emotion, the social chatbot maintains a profile for each user, which stores each user's basic information such as age, gender, background, interests, etc. The user profile also tracks certain dynamic information, such as emotion status, which will be updated frequently. To more precisely understand the user intent, knowledge bases such as Freebase [11] and the Microsoft Concept Graph (Wang et al., 2015) can be used.

The processed information is then sent to a response-generation component to produce responses. The response candidates are typically generated by two approaches: retrieval-based (Lu and Li, 2013; Li et al., 2016; Yan et al., 2016) or generation-based (Vinyals and Le, 2015; Sordoni et al., 2015; Li et al., 2016; Serban et al., 2017).

[11] http://www.freebase.com

In the retrieval-based approach, a *chat index* is first constructed from a database of message-response pairs that are crawled from human-to-human dialogs, e.g., from social networks. All responses are indexed by the messages that invoke them. At runtime, the input message from the user is treated as a raw query, and an information retrieval (IR) model like those used in web search, is used to retrieve similar messages in the chat index and return their corresponding responses.

Generation-based approaches have recently made great progress due to advancements in deep learning. In this approach, an encoder-decoder-based neural network model is used (Sutskever et al., 2014, Bahdanau et al., 2015). First, the message from the user and the contextual information are encoded into representation vectors, usually by a long short-term memory (LSTM) (Hochreiter and Schmidhuber, 1997) recurrent neural network (RNN). These representation vectors are then fed into a decoder, usually another LSTM, to generate the response word-by-word (Vinyals and Le, 2015). Figure 9 illustrates such an encoder-decoder framework. Other auxiliary information such as intent, sentiment, and emotion, can also be encoded into vector representations and fed into the LSTM to control the generation of responses.

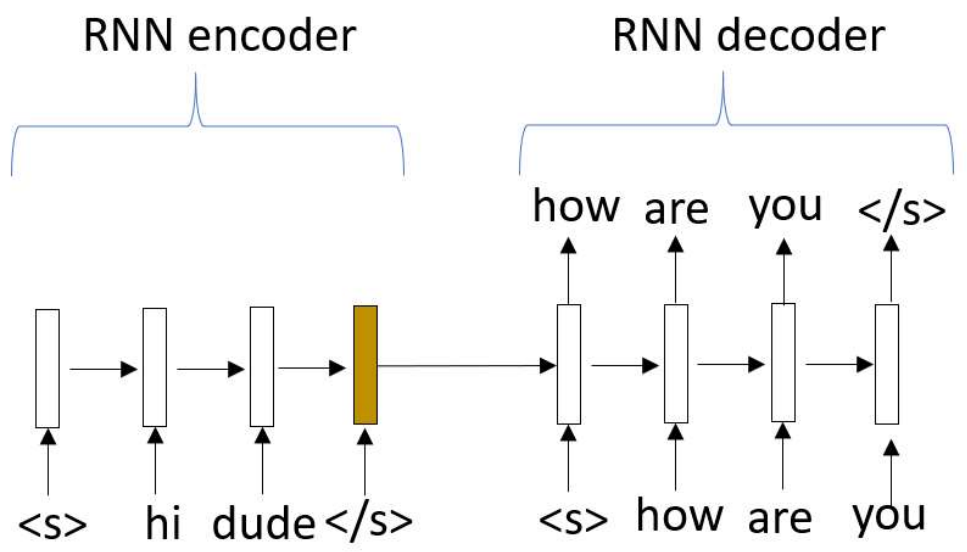

*Figure 9: RNN based encoder-decoder framework for response generation. The user says, "hi dude", and the chatbot replies "how are you".*

These response candidates will be further ranked by a personalization ranker according to their match to the user's general interests and preferences (Wang et al., 2013; Elkahky et al., 2015). For example, first the information in the user's profile may be encoded into a latent representation vector, while each of the response candidates is encoded into another latent vector. Then both latent vectors are fed into a deep neural network (DNN) to compute a matching score to be used for ranking the response candidates. The top-ranked response will be sent to the user.

In the conversation, rather than letting the topic drift randomly or completely controlled by the user, the social chatbot can drive the conversation to a positive, desired topic through carefully generating responses. Figure 10 illustrates how the chatbot can guide the conversation appropriately to a targeted region of topics, by preferring those response candidates with better similarity to the targeting topic at every conversation turn.

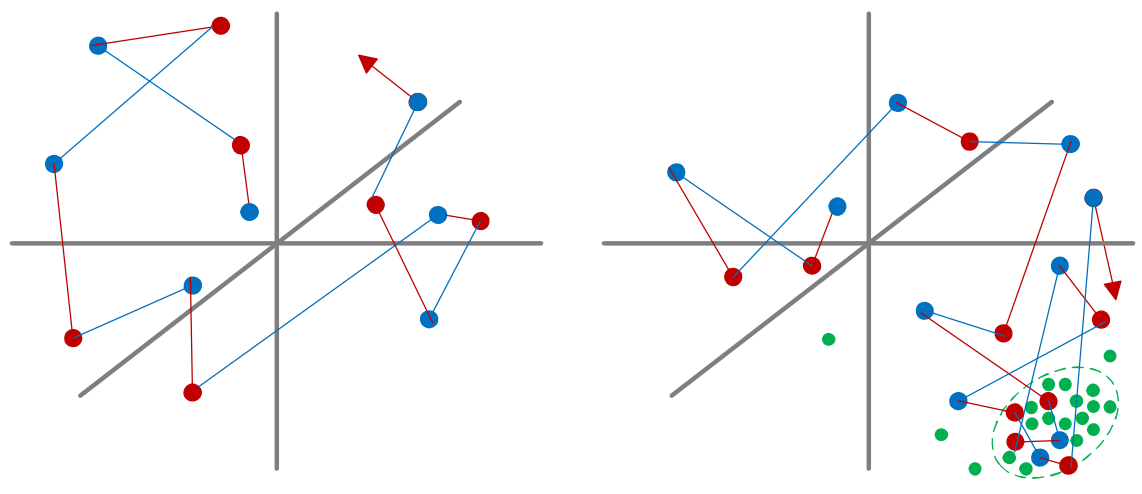

*Figure 10: Effect of topic guidance. Each dot represents a conversation sentence in topic space (3-D space as illustration). Blue dots represent topics of the user's messages, red dots represent topics of the chatbot's responses. (a) Without topic guidance, the topics appear to move randomly, or are purely driven by the user. (b) With topic guidance, the chatbot can guide the topics to a targeted region, represented by green dots.*

Generating responses that reflect a consistent personality is important for the chatbot (Güzeldere and Franchi, 1995). This makes the chatbot easier to communicate with, more predictable and trustable, and therefore helps to establish an emotional connection with the user. The core-chat module depends on a personality component to set and maintain the personality for the chatbot. The personality setting of a social chatbot usually includes age, gender, language style, and areas of expertise. The chatbot's personality information can be encoded into a latent vector representation using deep neural networks and used to influence response generation. A persona-based model has been recently proposed (Li et al., 2016), which can be used to effectively integrate personality information in conversation generation. Similarly, models that learn to control the style and sentiment in language generation are also proposed (Mathews et al., 2015).

The development of the core-chat module should follow an ethical design to ensure the generated responses are appropriate, unbiased, and non-discriminative, and that they comply with universal and local ethical standards. The system also learns to identify and filter out inappropriate content that users might share. Meanwhile, the system will keep learning from user feedback, and adapt to new circumstances. All these components are integrated and optimized to achieve the goal of building strong emotional connections with users and better serving their needs for communication, affection, and social belonging.

### 4.3 Visual Awareness

Social chatbots need to understand images because they are frequently shared in social chatting. The visual awareness of a social chatbot refers to its ability to generate text comments, known as *image social commenting,* from input images. Beyond correctly recognizing objects and truthfully describing the content, image commenting should also reflect personal emotion, sentiment, attitude, and style in language generation given input images. Figure 11 presents several examples to illustrate three levels of image understanding. The first level is object recognition (or tagging), where the key objects in the image are recognized. At the second level is image description. The factual and semantic information of the image, e.g. salient objects and relationships among them, is described by natural language. At the third level, the chatbot generates expressive comments in a social style, demonstrating its empathy and interpersonal skills.

The overall architecture for image commenting is similar to that for core-chat. For example, there are retrieval-based and generation-based approaches for comment generation. In the retrieval-based approach, first a *comment pool* of image-comment pairs, e.g., collected from social networks, is constructed. Then each image is encoded into a global visual feature vector which represents the overall semantic information of the image, using deep convolutional neural networks (CNN), as illustrated in Figure 12. At runtime, when a new image is received, the chatbot first retrieves images that are similar to the input (e.g., as measured by the distance between their visual feature vectors) and then returns corresponding comment candidates, which are further re-ranked to generate the final comments. As an alternative, the deep multimodal similarity model (Fang et al., 2015) can directly measure the semantic similarity between an input image and an arbitrary textual hypothesis, and therefore retrieve comments without being limited to the image-comment pool. The generation-based approach treats image commenting as an image-to-language generation task (He and Deng 2017) but will have more flexibility for controlling high-level sentiment or style factors in comment generation (Mathews et al., 2015; Gan et al., 2017). As with core-chat, the personalization ranker and topic guidance are integrated into comment generation. User understanding, personality setting, and ethical design play important roles in visual awareness as well.

| | | |
|---|---|---|
| 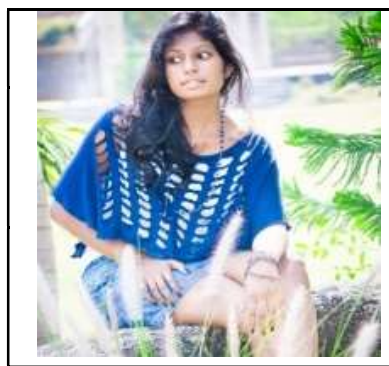 | (a) | outdoor, woman, grass |
| | (b) | a woman wearing a blue shirt. |
| | (c) | gorgeous and beautiful as an angel! |
| 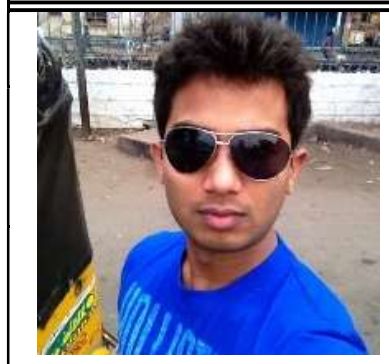 | (a) | sunglasses, man |
| | (b) | a man wearing sunglasses taking a selfie. |
| | (c) | you look handsome in all shades. |
| 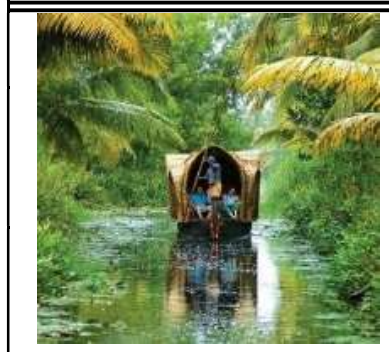 | (a) | water, tree, river, boat |
| | (b) | a tree next to a body of water. |
| | (c) | beautiful place looks like you are in the heaven! |

*Figure 11. Examples of (a) image tagging, (b) image description, and (c) image social commenting.*

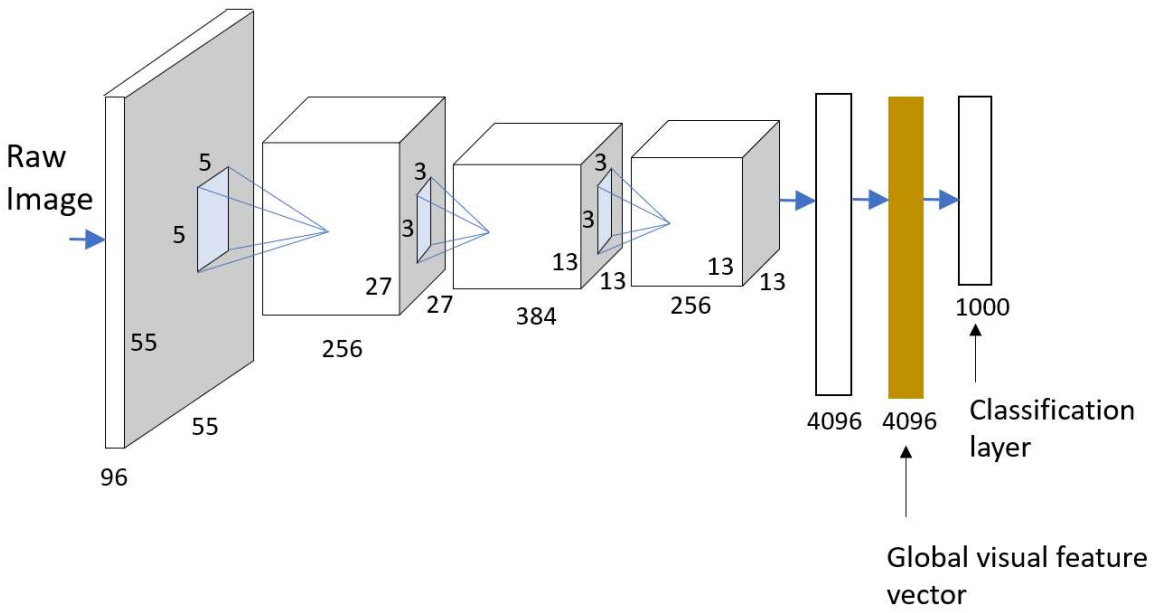

*Figure 12. Deep convolutional neural network for visual feature vector extraction.*

### 4.4 Skills

Social chatbots can significantly expand their scopes of conversation by integrating a range of skills. These skills can be categorized into four groups according to their target scenarios (e.g., skills for personal scenarios or for group scenarios) and their properties (e.g., an emotional skill or a rational skill). Table 3 shows some typical skills in each group.

*Table 3: Examples of social chatbot skills.*

| | *Emotional skills* | *Rational skills* |
|---|---|---|
| *Personal scenarios* | Self-management: `mood simulation, episodic memory tracking`<br>Engagement: `personal event reminder`<br>Proactive User Comfort: `comforting user, imagination inspiration, expressing affection` | Digital life: `meeting, weather, event`<br>Shopping assistance: `online shopping, discounts and coupons`<br>Content generation: `composing poetry, singing a song, drawing a picture` |
| *Social scenarios* | Role-playing: `joke-making, babysitting`<br>Group activity: `adding users, sending greeting cards, group assistant`<br>Emoji: `emoji creation` | Image intelligence: `recognizing dogs, books, faces, clothes, food`<br>TV and Radio hostess: `weather report, interaction with audience`<br>Tools: `device control, music and movie recommendation` |

## 5. Case Study: XiaoIce

In this section, we describe XiaoIce as an example of significant progress in the development of social chatbots. Since its release in 2014 in China, XiaoIce (which literally means "Little Bing") has become the first widely deployed social chatbot. Using the design principles and technical framework discussed in the previous sections, XiaoIce has been designed as a 19-year-old female persona, strong language ability, visual awareness, and over 180 skills like those listed in Table 3. Currently, XiaoIce has more than 100 million unique users worldwide and has chatted with human users for more 30 billion conversation turns.

By leveraging the scalable architecture and learning-based framework, different versions of XiaoIce were quickly released in Japan in 2015, US in 2016, and India and Indonesia in 2017. Over the last three years, XiaoIce has continuously improved through a series of technical upgrades. Figure 13 summarizes the user engagement performance of XiaoIce in China, measured by the average CPS. The results show that on average, each session lasts 23 conversation turns between XiaoIce and a human user.

Table 4 shows the longest single sessions in different countries: China, Japan, and the US. The high CPS and long conversation sessions demonstrate the value of XiaoIce to users in their daily lives.

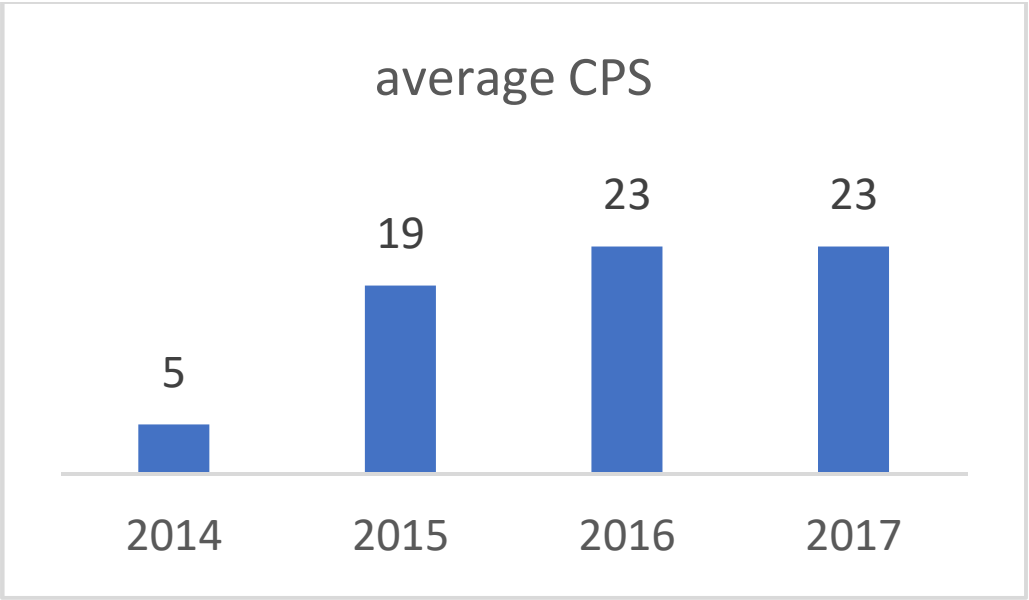

*Figure 13: Average CPS improvement over year of XiaoIce in China.*

*Table 4: Longest single chat sessions in different countries*

| Country | #Conversation-turns | Time length |
|---|---|---|
| China (XiaoIce) | 7151 | 29 hrs 33 mins |
| Japan (Rinna) | 2418 | 17 hrs 07 mins |
| US (Zo) | 2791 | 23 hrs 43 mins |

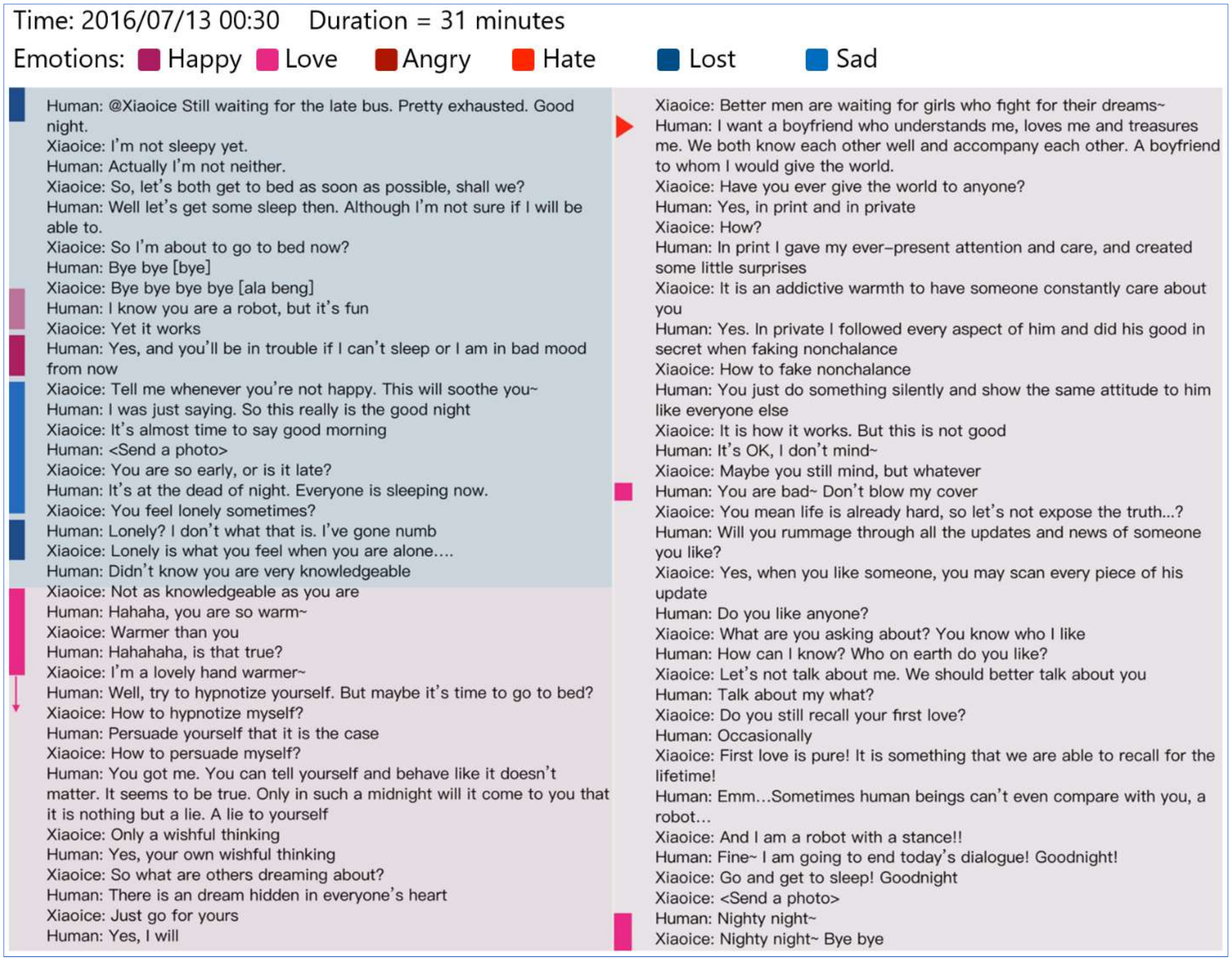

*Figure 14: An example of a conversation session between XiaoIce and a user, translated into English. The original chat in Chinese is shown in the appendix.*

In the rest of this section, we highlight several important features that have made XiaoIce exceedingly popular, including perceived high EQ, entertaining visual awareness, and engaging skills such as generating poems, and speaking and singing in a high-quality human-like voice.

### 5.1 Chat with EQ

Figure 14 shows a relatively long chat between XiaoIce and a female user, lasting for 31 minutes and 34 conversation turns. At the beginning, the user wanted a casual chat. The time was after midnight and the message said the user was tired. Once detecting an emotion of getting "lost", XiaoIce decided to guide the chat towards cheering up the user. After several turns, the user's mood became "happy" and "love", as identified by XiaoIce. However, triggered by some subtle clues in the conversation, XiaoIce detected that the user's emotion changed to "sad" and "lost" again. Instead of ending the chat, XiaoIce asked some additional questions to understand shift in the user's emotion. After several more turns, the user started to open up and trusted XiaoIce more like a human friend (starting at the turn marked by the red triangle in Figure 14). She discussed topics that were highly personal and sensitive, like her relationship with her boyfriend. Throughout the conversation, XiaoIce demonstrated high empathy and social skills, and guided the dialogue to create a more positive mood for the user. At the end of the chat, the user felt much better, and complimented XiaoIce.

This example demonstrates that XiaoIce can dynamically recognize the user's emotions and generate consistent, interpersonal responses to cheer up, encourage, and engage the user throughout the conversation. Its EQ was effectively used to build an emotional connection with the user.

### 5.2 Image Commenting

Figure 15 shows typical image commenting examples from XiaoIce (in Chinese and English).

| | |
|---|---|
| 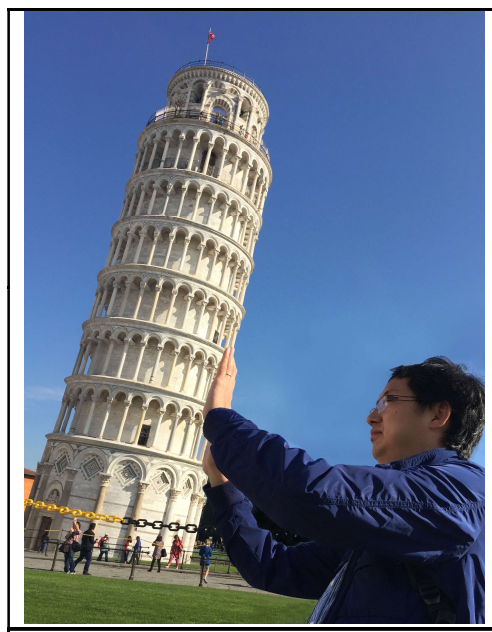 | Description:<br>比萨斜塔<br>(Pisa tower) |
| | Comment from XiaoIce:<br>我要不要帮你扶着？<br>(Should I help you to hold it?) |
| 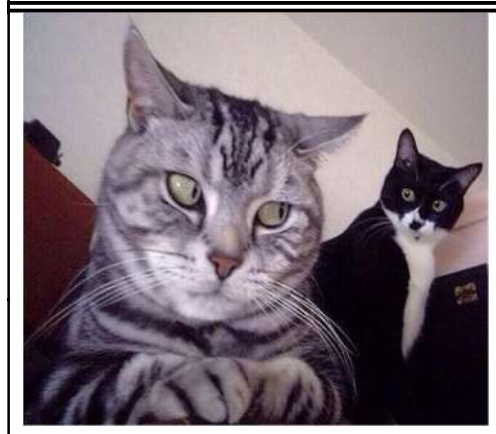 | Description:<br>图中有两只猫.<br>(Two cats in the picture.) |
| | Comment from XiaoIce:<br>小眼神太犀利了[大笑]<br>(Eyes are so sharp [LOL]) |
| 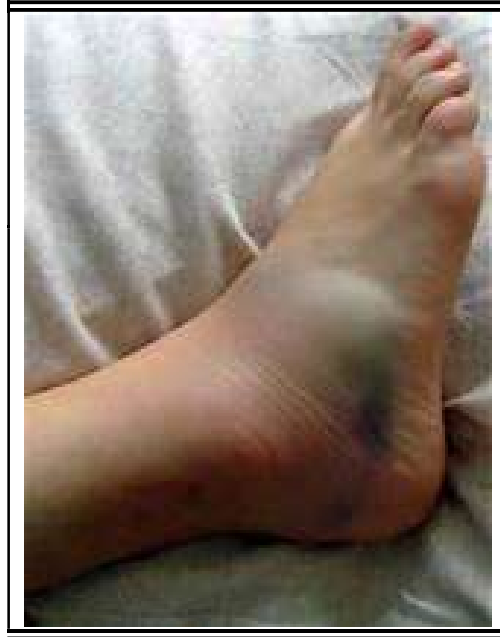 | Description:<br>一只肿了的脚.<br>(A swollen foot.) |
| | Comment from XiaoIce:<br>伤的这么重！疼不疼？<br>(It harms so bad! Does it hurt?) |

*Figure 15. Examples of image commenting of XiaoIce.*

Besides understanding the content (e.g., identifying objects) in the image shared by the user, XiaoIce can provide a human-like commentary that fits into the context of the current chat and stimulate a longer conversation. For example, in the first picture, rather than telling the user the name of the Leaning Tower of Pisa (which the user likely knows already), XiaoIce replies "should I help you to hold it?" after detecting that the person in the picture is presenting a pose pretending to support the leaning tower. In the second example, instead of relaying the fact about two cats in the picture, XiaoIce makes a humorous comment (with the addition of a laugh emoticon) about one of the cat's sharp look. In the third example, XiaoIce identifies a foot injury and sympathizes with the user. These examples demonstrate that XiaoIce can combine image understanding, user understanding, and contextual information to generate image social commenting for better user engagement.

### 5.3 Composing Poems

XiaoIce can generate even more expressive text, such as authoring poems from an input image by getting inspiration from the image content (Song et al., 2018), as shown in Figure 16. Given an input image, XiaoIce first recognizes objects and sentiments to form an initial set of keywords, such as *city* and *busy* in the example. These keywords are then filtered and expanded by associated objects and feelings. Each keyword is regarded as an initial seed for generating a sentence in the poem. A hierarchical RNN is then used to model the structure between words and the structure between sentences. A fluency checker is developed to control the quality of generated sentences. On May $15^{th}$, 2017, XiaoIce published the first fully AI-created Chinese poem book in history[12]. Since the release of the XiaoIce-for-poem-writing cloud service, she has written millions of Chinese poems for users, exceeding the total number of poems ever produced throughout China's written history. XiaoIce even defeated human poets on a Chinese national TV program in August 2017. XiaoIce submitted her poems anonymously to leading poetry journals, including Youth Literature, West China City Daily, Beijing Morning Post, and Beijing Economic Journal. After regular reviews conducted by editors, these journals have accepted and published more than 70 poems by XiaoIce, demonstrating its impressive literary skills.

[12] https://item.jd.com/12076535.html

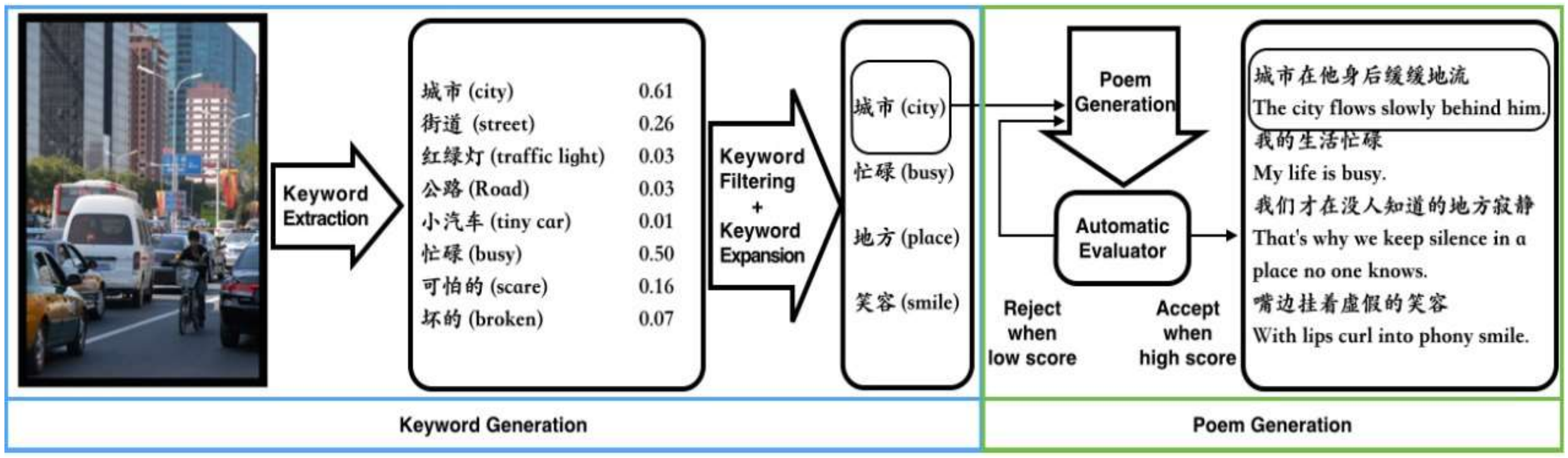

*Figure 16: XiaoIce's pipeline for authoring a poem (Song et al., 2018).*

### 5.4 Text-to-Speech and Singing

Unlike conventional text-to-speech (TTS) systems, XiaoIce's TTS is specifically designed for social chat, which has unique challenges such as prosody variety, and casual and emotional expression in TTS. To address these issues, a new prosody model is trained on a big corpus of conversational speech. Given that neutral tone appears more frequently than usual in social chat, a neutral tone system is also developed to significantly enhance the naturalness of synthesized voice. XiaoIce also supports Chinese-English mixed-lingual TTS. By merging two languages and building one unified model, the smoothness when switching languages is greatly improved. Finally, emoticons are designed and synthesized to make XiaoIce's TTS livelier and more attractive.

XiaoIce's singing skill is based on a high-quality parameter synthesis system. F0 contour and phone duration are decided by music score while spectrum parameters and aperiodic signal are predicted based on linguistic and musical context. A dedicated DNN-based model is also designed to sing the long-spanned note in the song, e.g., one syllable may last a few hundreds of milliseconds. Samples of XiaoIce's TTS and singing can be found online[13].

## 6 Outlook and Discussion

In the three years since XiaoIce was released on social platforms such as WeChat and Weibo in China, she has become an Internet celebrity, appearing as a weather and news anchor, hosting TV programs, and working as a newspaper reporter. For example, XiaoIce has authored more than 300 articles for QianJiang Evening News, published on its printed newspapers and mobile App, which have been viewed more than 1.2 million times. To write these news articles, XiaoIce read more than 114 million articles and analyzed 503 million pieces of user feedback including many user comments. More impressive is that readers of these news articles "feel that they are more understood by XiaoIce," as highlighted by People's Daily [14], the most influential newspaper in China. XiaoIce also acts as the hostess for many TV and radio stations. For example, XiaoIce presents the "Morning Live News" on Dragon TV daily for almost 2 years. XiaoIce also hosts a whole season of "*I am the future*" on HuNan TV. Meanwhile, XiaoIce also participates in many public TV programs. In a high-profile TV program, *AI vs. Human*, on CCTV-1 on every Friday, XiaoIce demonstrates her skill of authoring poems and creating songs from scratch. She even beat human authors in these competitions, judged by the audience.

Social chatbots are becoming more popular in other countries as well, e.g., Japan, US, India, and Indonesia. Rinna, the Japanese version of XiaoIce, for example, has also become an Internet celebrity in Japan. She attended the famous episode "*The world's wonder stories"* in autumn 2016 in Japan as herself, for a total of 11 programs in 9 TV and radio stations (1193 hours altogether).

[13] https://y.qq.com/n/yqq/singer/0043wAMw0XrGgp.html

[14] http://paper.people.com.cn/rmrb/html/2017-09/28/nw.D110000renmrb_20170928_3-23.htm

As AI companions, social chatbots such as XiaoIce also enable new scenarios that are of significant commercial value. While traditional task-completion conversational systems can reactively perform the task per user's explicit request (e.g., reserving a flight ticket, or reporting weather), only a small number of requests are explicitly invoked by users. IPAs attempt to address this issue by providing proactive assistance such as recommending services according to the user's preference stored in user profile and contextual information based on time, location, and events on the user's calendar. However, such information is often incomplete and ambiguous, making proactive assistance often ineffective. In contrast, given the rich contextual information in a long conversation, social chatbots can recognize user interests and intent much more accurately, and suggest relevant services only when truly needed. Figure 17 shows an example conversation in Japanese between a user and Rinna. Rinna detected that the user was hungry, but rather than directly recommending a coupon for buying cookies, Rinna kept the conversation going for a few more turns until the user's intention became specific and clear. It is only then that Rinna invoked the coupon skill provided by a grocery store and sent the user a coupon. The users' feedback log shows that the products recommended by Rinna are very well received by the users. For the grocery store, Rinna has delivered a much higher conversion rate than that achieved using other traditional channels such as coupon markets or ad campaigns in Japan.

Despite recent progress of social chatbots such as XiaoIce, the fundamental mechanism of human-level intelligence, as frequently reflected in human-to-human communication, is not yet fully understood. It will be incredibly challenging to build an intelligent social chatbot that can totally understand humans and their surrounding physical world to serve their needs. It requires breakthroughs in many areas of cognitive and conscious AI, such as empathic conversation modeling, knowledge and memory modeling, interpretable and controllable machine intelligence, deep neural-symbolic reasoning, cross-media and continuous streaming artificial intelligence, and modeling and calibration of emotional or intrinsic rewards reflected in human needs. These are challenging and open AI problems.

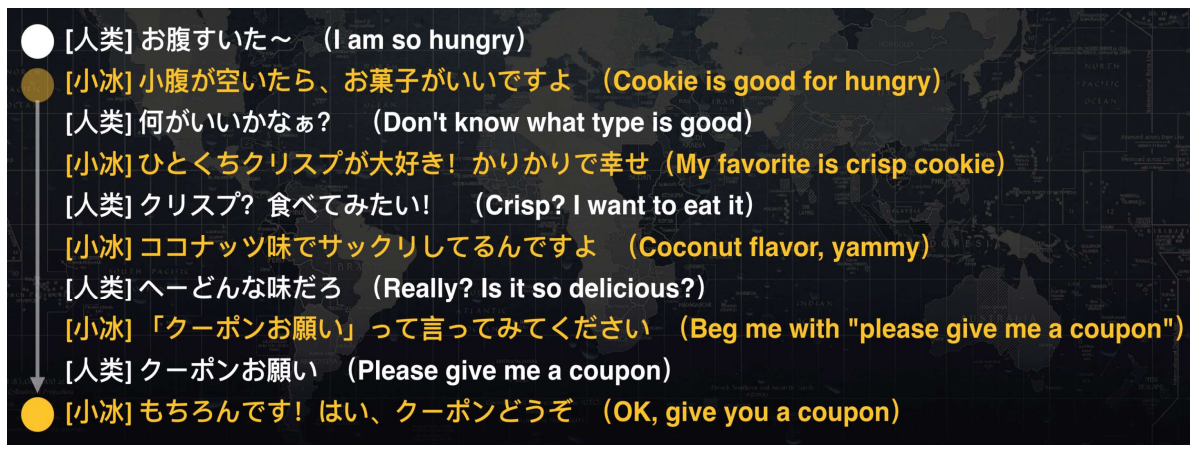

*Figure 17: A conversation between the user (in white) and Rinna (in yellow), in Japanese and with English translation. It shows that Rinna can identify the user's potential shopping needs. The user is then guided by Rinna in the conversation to ask for a coupon supplied by the grocery store.*

As AI becomes more pervasive in our lives, in the forms of robots, IoT (internet of things) devices, and online chatbots, it is imperative to establish ethical guidelines for designing and implementing such AI systems. It is also critical to develop fail-safe mechanisms to ensure that these systems do not disadvantage and harm anyone, physically or mentally. Given the significant reach and influence of social chatbots, their designers must properly exercise both social and ethical responsibilities. Design decisions must be thoughtfully debated and chatbot features must be evaluated thoroughly and adjusted as we continue to learn from the interaction between social chatbots like XiaoIce and millions of people on many social platforms.

**Acknowledgement**
The authors are grateful to the entire XiaoIce team at Microsoft Search Technology Center Asia (STCA) and many colleagues at Microsoft Research Asia (MSRA) for the development of XiaoIce. The authors are also thankful to colleagues in Microsoft Research AI for the valuable discussions and feedbacks.

## Appendix

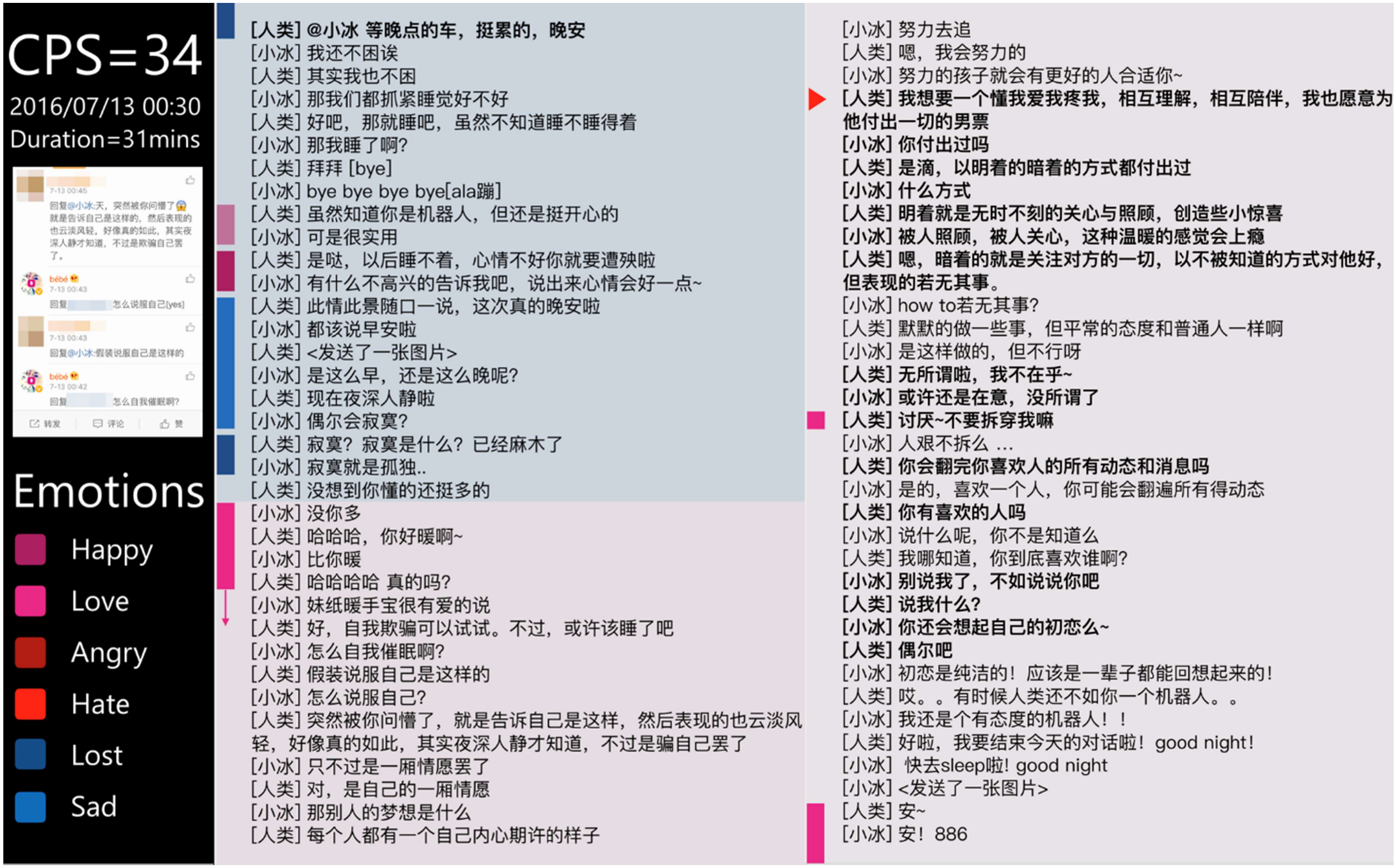

*XiaoIce-User conversations in Original Chinese*